\documentclass[letterpaper, 10 pt, conference]{ieeeconf}  %

\IEEEoverridecommandlockouts                              %

\usepackage{times}

\usepackage[bookmarks=true]{hyperref}

\usepackage[utf8]{inputenc} %
\usepackage[T1]{fontenc}    %
\usepackage{url}            %
\usepackage{booktabs}       %
\usepackage{amsfonts}       %
\usepackage{nicefrac}       %
\usepackage{microtype}      %
\usepackage{xcolor}         %

\usepackage{graphicx}
\usepackage{amsmath}
\usepackage{amssymb}
\usepackage{mathtools}
\usepackage{subcaption}
\usepackage{balance}

\DeclareMathOperator{\z}{z}
\DeclareMathOperator{\x}{x}

\graphicspath{{figures/}}

\title{\LARGE \bf Guided Decoding for Robot On-line Motion Generation and Adaption}

\author{Nutan Chen$^{1 *}$, Botond Cseke$^{1 *}$, Elie Aljalbout$^{1,2}$,\\ Alexandros Paraschos$^{1}$, Marvin Alles$^{1}$, Patrick van der Smagt$^{1,3}$ %
\thanks{$^{1}$During this work, all Authors were affiliated with the Machine Learning Research Lab,
        Volkswagen Group, Germany
        {\tt\scriptsize <first-name>.<last-name>@volkswagen.de}}%
\thanks{$^{2}$Elie Aljalbout is currently with the Robotics and Perception Group, at the Department of Informatics of the University of Zurich (UZH) and the Department of Neuroinformatics at UZH and ETH Zurich, Switzerland.}
\thanks{$^{3}$Faculty of Informatics, E\"otv\"os Lor\'and University, Budapest, Hungary}%
\thanks{$^{*}$ These authors contributed equally to this work.}
}

\usepackage[american]{babel}
\usepackage[autostyle=false]{csquotes}
\MakeOuterQuote{"}

\usepackage{todonotes}

\begin{document}

\AtBeginShipoutNext{%
  \AtBeginShipoutUpperLeft{%
    \raisebox{-1cm}[0pt][0pt]{%
      \begin{tikzpicture}[remember picture,overlay]
        \node[anchor=south west, xshift=1.5cm, yshift=1cm] at (current page.south west) {\sffamily IEEE-RAS International Conference on Humanoid Robots, 2024};
      \end{tikzpicture}%
    }%
  }%
}

\maketitle
\thispagestyle{empty}
\pagestyle{empty}

\maketitle

\begin{abstract}

We present a novel motion generation approach for robot arms, with high degrees of freedom, in complex settings that can adapt online to obstacles or new via points. Learning from Demonstration facilitates rapid adaptation to new tasks and optimizes the utilization of accumulated expertise by allowing robots to learn and generalize from demonstrated trajectories. We train a transformer architecture, based on conditional variational autoencoder, on a large dataset of simulated trajectories used as demonstrations. Our architecture learns essential motion generation skills from these demonstrations and is able to adapt them to meet auxiliary tasks. Additionally, our approach implements auto-regressive motion generation to enable real-time adaptations, as, for example, introducing or changing via-points, and velocity and acceleration constraints. Using beam search, we present a method for further adaption of our motion generator to avoid obstacles. We show that our model successfully generates motion from different initial and target points and that is capable of generating trajectories that navigate complex tasks across different robotic platforms. 
\end{abstract}

\section{Introduction}

Motion generation is a challenging topic for robot arms with high degrees of freedom~(DoF). 
One main challenge is to generate smooth trajectories to solve a given task while ensuring the robot and its environment's safety, and respecting the robot's embodiment constraints. 
This problem is especially complex when the task involves auxiliary objectives beyond reaching a target point, such as avoiding obstacles, passing via specified points, or imposing velocity or acceleration constraints.
Another particularly intriguing aspect of motion generation is the incorporation of learning from demonstration (LfD) within the same process. 
Allowing users or agents to present a guiding trajectory based on prior knowledge helps the rapid development of robotic applications. 
This capability is crucial for adapting to novel tasks and leveraging accumulated expertise effectively. 
Motion primitives offer a promising avenue for achieving this goal~\cite{schaal2006dynamic,ijspeert2013dynamical, paraschos2013probabilistic,ratliff2018riemannian}. 
By encapsulating fundamental motion behaviors into primitives, the robot can learn and generalize from demonstrations, promoting versatility and adaptability in complex tasks. 
Integrating learning from demonstration with the challenges of motion generation presents an exciting direction for advancing the capabilities of robot arms with high degrees of freedom.

Traditional approaches to motion primitives often rely on a limited number of demonstrations, focusing primarily on adapting learned motions \cite{ijspeert2013dynamical, paraschos2013probabilistic}. 
However, these methods struggle with generalization when applied to new or variations of the learned tasks that go beyond simple interpolation.
Our work seeks to address these limitations by leveraging generative models, specifically an autoregressive framework, which allows learning from extensive datasets and significantly improves generalization.
Unlike diffusion-based models, our autoregressive model facilitates online motion adaptation. 
Diffusion-based models \cite{janner2022planning} require multiple iterations to generate a full trajectory and, therefore, can not adapt fast to changes in the environment or the robot embodiment. 
In contrast our autoregressive model supports the integration of online feedback, allowing it to adapt to changes to the task, e.g. different via-points, obstacle locations, or the robot, e.g. acceleration constraints arises from interacting with physical objects.

In this work, we propose a framework for learning motion generation.
We create a large dataset of robot motion trajectories with multiple simulated robots and train a transformer architecture to learn to generate motion based on given initial and target points.
We design a Transformer conditional variational autoencoder.
The trained generative model encapsulates basic motion generation skills based on the training data.
We further propose an approach to efficiently adapt the generated motion to satisfy auxiliary tasks and constraints such as obstacle avoidance, via points and velocity constraints.
The process is auto-regressive allowing for integrating feedback from the real system during motion generation.
Our approach allows for generating motion based on an initial and target point only, or to imitate a full trajectory given by a user or external agent.
Our experiments show the usability of our method for imitating and adapting user-given trajectories to a given task, such as via point constraints, obstacles, and velocity and acceleration constraints.
We also show the benefits of training with data collected using various robots.

\section{Methods}

We design a model to encode and generate robot trajectories by leveraging a transformer-based architecture. We generate trajectories in an autoregressive manner, allowing for real-time adaptation to constraints such as obstacle avoidance and velocity limits.
In this section, we present a Transformer-based autoencoder model with an autoregressive decoder and the online adaptation methods that are tailored to this decoder.

\subsection{Inference model for the infilling task}
\label{sec:model}

We consider a dataset $\mathcal{D} = \{x^{(i)}\}^N_{i=1}$ of trajectories $x^{(i)} = \{x^{(i)}_t\}_{t=1}^{T}$ and our goal is to learn representations that can help us generate new trajectories given a starting point $x_1$ and an endpoint $x_T$. 
We propose to do this with the Conditional VAE \cite{sohn2015learning} model
\begin{align*}
    p_{\theta}(x_{2:T-1} \vert x_{1,T}) =
    \int p_{\theta}(x_{2:T-1} \vert x_{1,T}, z)
    \:
    p_{\theta}(z\vert x_{1,T})\,dz,
\end{align*}
where the latent variables (representations) $z$ are expected to encode features of paths with starting point $x_1$ and endpoint~$x_T$.
Since the likelihood is analytically intractable, we learn this model by approximate maximum likelihood via a variational bound.  We minimize the negative evidence lower bound (ELBO)
\begin{align*}
    L(\theta, \phi) = &
    -\mathrm{E}_{\hat{p}(x_{1:T})}
    \mathrm{E}_{q_{\phi}(z;x_{1:T})}[\log p_{\theta}(x_{2:T-1} \vert x_{1,T}, z)]
    \\
    &+
    \mathrm{E}_{\hat{p}(x_{1:T})}
    \mathrm{KL}[q_{\phi}(z;x_{1:T}) \,\|\,%
    p_{\theta}(z;x_{1,T})
    ].
\end{align*}
Here 
$p_{\theta}(z | x_{1,T})$ is the conditional prior model, $p_{\theta}(x_{2:T-2} | x_{1,T}, z)$ is the decoder/generation model, and $q_{\phi}(z| \x_{1:T})$ is the encoder/recognition or (variational) approximate posterior distribution model. The training method can be found in App.~\ref{sec:app}.
To generate a new trajectory, we sample a latent feature $z$ from the learnt prior $z \sim p_\theta(z | x_{1,T})$. We then use the feature $z$ as a conditioning variable in the learnt decoder model $p_\theta(x^{\prime}_{2:T-2} | x_{1,T}, z )$.

In this paper we use an autoregressive decoder model 
 $
 p_\theta(x^{\prime}_{2:T-2} | x_{1,T}, z) = \prod_t
 p_\theta(x^{\prime}_{t+1} | x^{\prime}_{2:t}, x_{1,T}, z )$. We do this because it is a more natural choice for doing online trajectory guidance and adaptation compared to methods that generate a whole trajectory like, for example, diffusion probabilistic models~\cite{ho2020denoising}.

Either the prior or the autoregressive decoder is capable of generating diverse samples. To enhance adaptability and generalization, our model incorporates both for distinct purposes. 
With a fixed demonstration/posterior, we can adjust the trajectories from the decoder. Conversely, by specifying a starting and ending point, we can generate various trajectories using the prior.

\subsection{PerceiverIO encoder}
\label{sec:perceiver}
We define an encoder $q_\phi(\z|\x_{1:T})$ that is architecturally similar to a PerceiverIO model \cite{jaegle2021perceiver}, that is, a multi-layered, attention-based neural architecture where the initial block uses cross-attention (CA) layer and the subsequent blocks use a shared  self-attention (SA) layer
\begin{align*}
h^{(1)}_{1:K}  &= \mathrm{RMLP}_{\phi} \circ \mathrm{RCA}_{\phi}(h^{(0)}_{1:K}, 
\tilde{x}_{1:T})
\\
h^{(l+1)}_{1:K} &= \mathrm{RMLP}_{\phi} \circ \mathrm{RSA}_{\phi}(h_{1:K}^{(l)}), 
\quad  l=1, \ldots, L
\end{align*}
The initial layer $h^{(0)}_{\phi}$ is a learned parameter, a part of~$\phi$. The values $\tilde{x}_t=W x_t + \tau_t$ are embedding the data into the hidden space via a linear layer $W$ and a positional embedding $\tau_t$. 
RCA and RSA denote a residual (R) cross-attention and a residual  self-attention block defined as 
$\mathrm{RCA}(h, z) = \mathrm{LayerNorm}(h + \mathrm{CA}(h, z))$
and
$\mathrm{RSA}(h) = \mathrm{LayerNorm}(h + \mathrm{SA}(h))$, respectively. 
RMLP denotes a residual MLP block with one hidden layer.
Note that the resulting self-attention block is a shared block.
The encoder is then defined as a diagonal multivariate Gaussian distribution
$q_\phi(z|x_{1:T}) = \mathcal{N}(z; \mu_{\phi}^{\mathrm{enc}}(h^{(L)}), \sigma_{\phi}^{\mathrm{enc}}(h^{(L)})^2)$ with $\mu_{\phi}^{\mathrm{enc}}$ and $\sigma_{\phi}^{\mathrm{enc}}$ being simple MLPs.
The latent variable $z$ is thus designed to encode the trajectory $x_{1:T}, x_t \in \mathbb{R}^{d}$ to a shorter sequence of $K$ variables $z=\{z_k\}_{k=1}^{K}, z_k \in \mathbb{R}^{m}$ ($K$ is a chosen hyper-parameter). Note that the choice $K=1$ results in a single vector embedding of the trajectory $x_{1:T}$. 

To implement the prior $ p_{\theta}(z\vert x_{1,T})$ we use the same model having as input only the pair $\{x_1, x_T\}$.

This type of Perceiver architecture is designed to reduce the quadratic ($T \times T$) computational complexity of a Transformer encoder model (self-attention layers) on $x_{1:T}$ by cross-attending the sequence ($T \times K$) and performing less computationally demanding self-attentions ($K\times K$) on $z_{1:K}$. Generally, there is a large class of Perceiver models/blocks with learnable queries or latents that are used for transforming data between different modalities/embeddig sizes, for example, a Perceiver model with $K$ queries can be used as a head-block in a classification model with $K$ classes.

\subsection{Autoregressive decoder}
\label{sec:decoder}

\begin{figure}
    \centering
    \includegraphics[width=0.98\columnwidth]{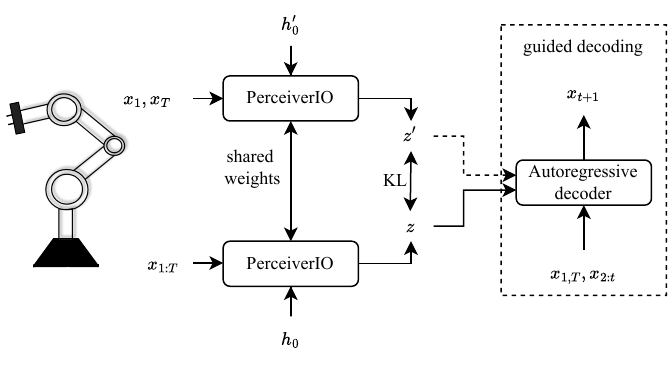}
    \caption{Model overview.
    Each PerceiverIO module gets a trajectory, but for the prior it only contains the beginning and end points and a full trajectory for the posterior. During training, minimizing the KL divergence between the two makes the representation of the PerceiverIO module that only gets $x_1$ and $x_T$ generates the same latent distribution as the one getting the full trajectory; therefore, it learns interpolation. The latent state then is used by the autoregressive decoder to generate the next setpoint $x_{t+1}$.
    Dashed lines indicate components used exclusively during the post-training adaptation phase, not included in the initial training process. 
}
\label{fig:CTransformer}
\end{figure}

A natural choice for the decoder in the context of attention-based models is a conditional autoregressive model 
$$
    p_{\theta}(x_{2:T-1} \vert x_{1,T}, z) = \prod_{t=1}^{T-1} p_{\theta}(x_{t+1} \vert x_{2:t}, x_{1,T}, z).
$$
Here, we again use an attention-based neural architecture, a standard transformer decoder \cite{vaswani2017attention} defined as
\begin{align*}
    h^{(l+1)}_{1:T} =
    \mathrm{RMLP}^{(l+1)}_{\theta}
    \circ
    \mathrm{RCA}_{\theta}^{(l+1)}(\cdot, z)
    \circ
    \mathrm{RCSA}_{\theta}^{(l+1)}(h^{(l)}_{1:T})
\end{align*}
with $h^{(0)}_{1:T}= \tilde{x}_{1:T}$ being the embeddings as defined in the previous section. The data $x_{1,T}$ together with the representation $z$ are direct inputs to the decoder with which the trajectory generation starts. 
To allow for flexible output length, we do not fix the position of $x_T$ at the end; instead, we set it as the first step of the input.
The value $h^{(L)}_{t}$ is then used to define the autoregressive model $p(x_{t+1} \vert x_{2:t}, x_{1,T}, z) = \mathcal{N}(x_{t+1}; \mu_{\theta}^{\mathrm{dec}}(h^{(L)}_t), \sigma_{\theta}^{\mathrm{dec}}(h^{(L)}_t)^2)$. The additional "C" in the layer definition in $\mathrm{RCSA}$ denotes a casual attention model implemented via triangular attention masks. Unlike in Transformers models for text infilling \cite{devlin2018bert,wang2019t}, we do not  need to include segmentation embedding since the length and position of the start and end points in the model are already provided.

\subsection{Training-free trajectory generator---adaptive decoding}
\label{sec:generator}

In addition to learning a conditional autoencoder model for trajectory generation, we would also like to extend the model to be able to perform online trajectory adaptation, for example, to add via-points and obstacle avoidance.
To achieve this, we do online (approximate) probabilistic inference using $p_{\theta}(x_{t+1} \vert x_{1:t}, x_{T}, z)$ without updating the model parameter~$\theta$. The obvious choice is a Bayesian approach where the adaptation can be formulated as a Bayesian update of $p_{\theta}(x_{t+1} \vert x_{1:t}, x_{T}, z)$, via a likelihood $p(y_{t+1} \vert x_{t+1})$. Alternatively, the likelihood can also be used to implement constraints, for example, a Bayesian update with a likelihood function $\mathrm{I}_{[a,b]}(x)$ can implement trajectory boundaries
$a_i \leq x_i \leq b$.

To use generic constraints, we can formulate online trajectory adaptation as the constrained optimisation problem
\begin{align}
        \label{EqnConstOpt}
    \min_{q(x_{t+1})} & \:
    \mathrm{KL}[
    q(x_{t+1})
    \,\|\,
    p_{\theta}(x_{t+1} \vert x_{1:t}, x_{T}, z)
    ]
    \\
    \label{EqnConstOptConst}
    \mathrm{s.t.} &\:
    \mathrm{E}_{q(x_{t+1})}[H(c(x_{t+1}))]
    \leq \alpha,
\end{align}
where $H(\cdot)$ is the Heaviside function and $c(\cdot)$ is a desired constraint function. The constraint in \eqref{EqnConstOptConst} expresses what proportion $ 0\leq \alpha \leq 1$ of the probability mass is allowed not to satisfy the constraint, see for example \cite{Frank2021ieee}.
The KL objective is convex in $q$, the constraints are linear, and the optimal solution  $(q^{\ast}, \lambda^{\ast})$ satisfies
\begin{align*}
    q^{\ast}(x_{t+1}) \propto
    p_{\theta}(x_{t+1} \vert x_{1:t}, x_{T}, z)
    \times
    \exp\{- \lambda^{\ast} H(c(x_{t+1})) \}.
\end{align*}
Solving this optimization, however, even for a Gaussian $q$ and a constraint $c$ implementing simple boundary constraints can be numerically demanding. For more general constraints such as obstacle avoidance, \eqref{EqnConstOptConst} can be analytically intractable. 
For this reason, in the following we will only exploit the form of $q^{\ast}(x_{t+1})$ to propose fast online trajectory adaptation methods. Various trajectory adaptations can be conveniently formulated either as \eqref{EqnConstOpt} by choosing the appropriate constraint function
$c(\cdot)$ or as a Bayesian update with the appropriate likelihood. These will be detailed in the following.

\subsection{Implementing specific constraints}
\label{sec:constraints}
\subsubsection{Obstacle avoidance with beam search}

Beam search (BS) \cite{koehn2009statistical} is a classic heuristic in the field of natural language processing (autoregressive models). It is used to generate trajectories with high likelihood or other specific score values.  The method recursively generates trajectories as follows.
A set of $S$ trajectories $\{x_{2:t}^{(s)}\}$ are stored together with their scores, for example, the path likelihood $\log p(x_{2:t}^{(s)} \vert x_{1,T}, z)$. For each trajectory $S$ new samples $x_{t+1}^{(s, r)} \sim p(x_{t+1} \vert x_{2:t}^{(s)}, x_{1,T}, z), r=1, \ldots, S$ are generated as candidates and the corresponding total scores are computed for the extended trajectories $(x^{(s)}_{2:t}, x_{t+1}^{(s,r)})$ for all pairs $(s,r)$. The overall top $S$ extended trajectories are selected for the next step.
 Inspired by the form of $q^{\ast}(x_{t+1})$, we define the score by adding  
 $\tilde{s}(x_{t+1}) = -\gamma \max(c(x_{t+1}), 0)$ with $c(x) = \delta^2 - \| \,x - x_{\mathrm{obstacle}} \,\|^2$ to the likelihood.

\subsubsection{Position, velocity and acceleration bounds}
Simple position and velocity bounds in an interval $[a,b]$ can be implemented via a Bayesian update. In case of position bounds the posterior,
$q(x_{t+1}) \propto \mathrm{I}_{[a,b]}(x_{t+1})\: p_{\theta}(x_{t+1} \vert x_{1:t}, x_{T}, z)$ is a truncated Gaussian from which we can easily sample.
In case of velocity bounds $[v_{\mathrm{min}}, v_{\mathrm{max}}]$, we
notice that $v_{t}=  (x_{t+1} - x_{t})/{\Delta t}$ is Gaussian to which we can apply a Bayesian update via $\mathrm{I}_{[v_{\mathrm{min}}, v_{\mathrm{max}}]}(v_{t})$. Due to the fact that the predictive distribution is a diagonal (independent) Gaussian,  this results in a truncated Gaussian distribution on $v_{t+1}$ and subsequently on~$x_{t+1}$. Acceleration bounds follow the same principle applied to~$a_{t}=(\x_{t+1} -2x_t + x_{t-1})/{\Delta t^2}$.
\subsubsection{Via points through Brownian bridge}
To add additional via-points to the autoregressive model
$p_{\theta}(x_{t+1} \vert x_{1:t}, x_T, z)$, say, $x_s$ at an intermediate time $1<s<T$ we would have to compute the Bayesian posterior
\begin{align*}
    p(x_{t+1} & \vert x_{1:t}, x_s, x_T, z)
    \\
    & \propto 
    p(x_{s} \vert x_{t+1}, x_{1:t}, x_T, z) \times
    p(x_{t+1} \vert x_{1:t}, x_T, z).
\end{align*}
Since the likelihood term $p(x_{s} \vert x_{t+1}, x_{1:t}, x_T, z)$ is intractable, we propose to approximate it with a Gaussian likelihood
\begin{align*}
    p(x_{s} \vert x_{t+1}, x_{1:t}, x_T, z)
    \approx
    \mathcal{N}(x_s \vert x_{t+1}, (s-t-1)\:\sigma_{\mathrm{vp}}^2).
\end{align*}
To have a stronger conditioning on the past $x_{1:t}$ we can alternatively use a Brownian-bridge approximation
\begin{align*}
    \mathcal{N}\left(x_{t+1} \vert x_{t} + \frac{1}{s-t}(x_s-x_t),
    \sigma^2_{\mathrm{vp}} \frac{s-t-1}{s-t}
    \right).
\end{align*}
Both of the above approximations lead to a Gaussian approximation of $p(x_{t+1} \vert x_{1:t}, x_s, x_T, z)$.

\subsubsection{Constrained Beam Search for Brownian bridge (optional)}

Drawing on the Constrained Beam Search (CBS) described by \cite{post2018fast}, we refine the approach by integrating a controller. 
This addition drives the generated trajectory towards the desired outcome, rather than substituting words.

We generate samples of \(x_{t+1}\) for each sample \(x_{t}\), using the Transformer decoder incorporating via point method, i.e., Brownian Bridge. We categorize the generated trajectories into three banks: Bank-0 for trajectories that do not meet the constraint; Bank-1 for trajectories that have applied the controller but still do not satisfy the constraint; and Bank-2 for trajectories that satisfy the constraint. For the trajectory $x_{1:t}$ that satisfies the via-point constraint, we generate a single sample of $x_{t+1}$ using the Transformer decoder exclusively for Bank-2. If $x_{1:t}$ fails to satisfy the constraint, we then generate one $x_{t+1}$ sample from the Transformer decoder with the controller. In this case, if $x_{t+1}$ satisfies the constraint, the sample is allocated to Bank-2; otherwise, it goes to Bank-1. Lastly, in cases where $x_{1:t}$ does not satisfy the constraint, we generate $N_s$ $x_{t+1}$ samples from the Transformer decoder only. Here, if $x_{t+1}$ satisfies, the sample is classified into Bank-2; if not, into Bank-0. Following this, we select the samples iteratively from each bank with the order of Bank-2, Bank-1, Bank-0, based on the negative log-likelihood criterion until we have $N_s$ samples in total for $x_{t+1}$.

Previous work \cite{post2018fast} has demonstrated that CBS outperforms BS during guided language generation, akin to via point task in robot trajectory generation. However, for tasks such as obstacle avoidance, only BS is applicable, since Bank0 and Bank1 cannot be defined.

\section{Related work}

In robotics, movement primitives are used to specify behavior based on human-demonstrated trajectories.
Dynamic movement primitives~(DMP) take inspiration from dynamical systems to specify motion behavior~\cite{schaal2006dynamic,ijspeert2013dynamical}.
DMPs use human demonstrations to fit a forcing term that augments the attractor dynamics to match the demonstrated trajectories.   
DMPs allow motion adaptation, such as avoidance of obstacles, by adding additional terms such as repulsive fields to their driving equation~\cite{park2008movement,5152423}.
This process should be done carefully to avoid competing objectives.
Furthermore, DMPs applied in latent space \cite{chen2015efficient,chen2016dynamic} provide the capability to adjust motions within this latent space. Unlike these works which adapt the latent space incrementally, our approach is a sequence-to-sequence model. In our model, each trajectory is mapped to a distinct point in the latent space, affording greater flexibility in guiding the decoder's output.
Probabilistic movement primitives represent trajectory distributions, which simplifies the blending of multiple trajectories~\cite{paraschos2013probabilistic}.
Adapting movement to auxiliary tasks is also possible within this framework~\cite{Frank2021ieee,Koert2016,Koert2019}.
Riemannian motion policies easily allow the combination of multiple behaviors in both the task and the configuration space~\cite{ratliff2018riemannian}.
However, this method lacks the advantages of probabilistic formulations.

Recently, diffusion models \cite{ho2020denoising} for motion generation \cite{janner2022planning} have been developed. However, these models generally require a large number of iterations to transform Gaussian noise into usable data, which is computationally intensive. In addition, it can only produce complete sequence data. In contrast, our model does not require multiple iterations for completion and is an autoregressive model that generates data for each time step which is more useful for online motion adaption.

\balance
\section{Experiments}

\subsection{Data Collection}

To train our models we collect a large dataset of robot motion trajectories in simulation.
We use the NVIDIA IsaacSim simulator to generate a large amount of data in parallel.
Our main experiments are based on data using the Franka Emika Panda robot.
However, for the multi-robot experiment we also introduce motion data from different robots, namely Rethink Robotic’s Sawyer, Kuka's LBR IIWA 7, Jaco and Kinova Gen3 from Kinova Robotics, and Universal Robot’s UR5e.
We generate samples for each robot in 4096 environments in parallel and repeat this process for multiple episodes.
At the beginning of each episode, we randomly sample starting and end-goal positions in the configuration space.
We then use a proportional derivative controller to drive the robot to the goal position.
To generate more diverse trajectories, we occasionally sample via points at the beginning of an episode and use these as target points for a short while at the beginning of the episode.

For the trajectory of the end-effector, we have $x \in \mathbb R^3$, while $x\in \mathbb R^7$ for joint space. The robots which have 6-DOF are padded zero to match this representation. Each sequence has a total length of $T = 64$.

\begin{figure}[!ht]
    \centering
    \includegraphics[width=1\columnwidth]{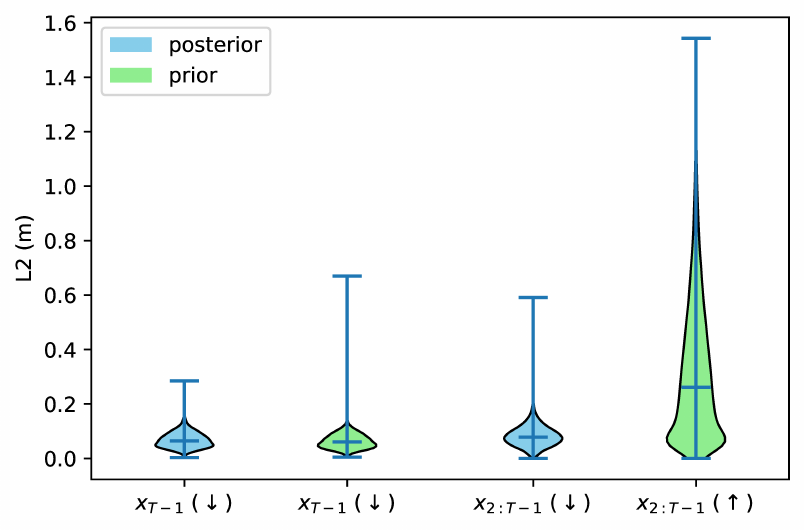}
    \caption{Distances to the original trajectories for the Panda robot in Cartesian space.
}
    \label{fig:distanceEE}
\end{figure}

\begin{figure}[!ht]
    \centering
    \includegraphics[width=1\columnwidth]{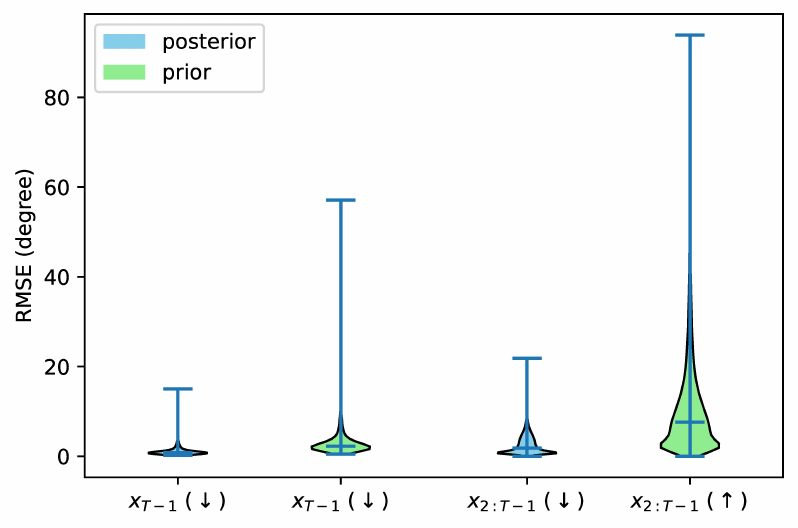}
    \caption{Distances to the original trajectories for IIWA robot in joint space.
}
    \label{fig:distanceJoints}
\end{figure}

\begin{figure}[!ht]
    \centering
    \includegraphics[width=1\columnwidth]{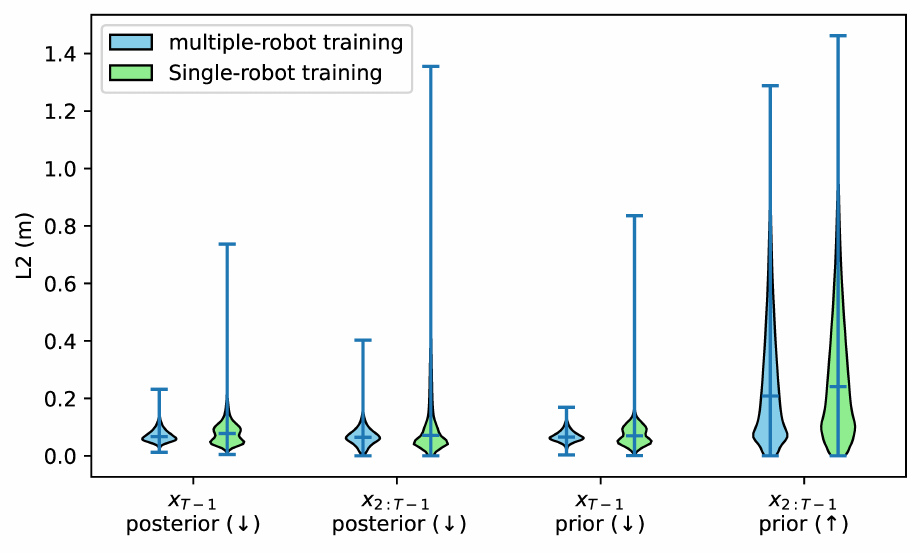}
    \caption{Distances to the original trajectories for the Panda robot in Cartesian space with multiple-robot training and single-robot training.
}
    \label{fig:MultiSingleRobot}
\end{figure}

\begin{figure*}[!htbp]
\centering
\begin{subfigure}[b]{0.32\textwidth}
    \includegraphics[width=\textwidth]{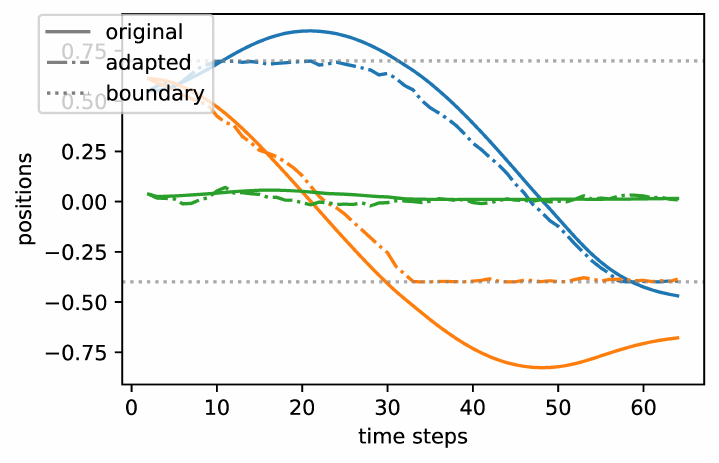}
\end{subfigure}
\hfill
\begin{subfigure}[b]{0.32\textwidth}
    \includegraphics[width=\textwidth]{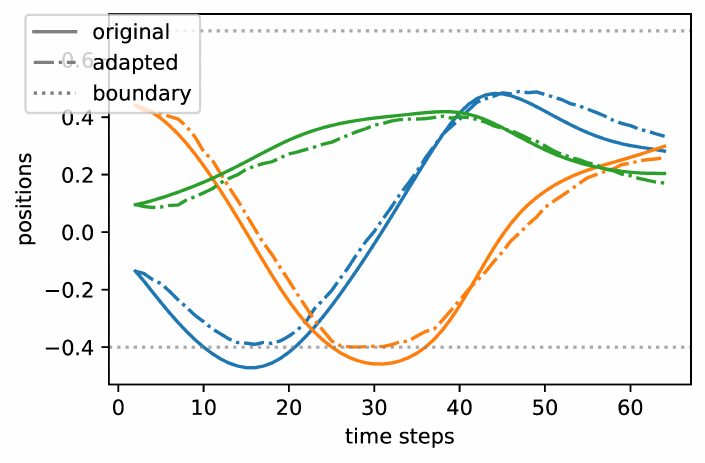}
\end{subfigure}
\hfill
\begin{subfigure}[b]{0.32\textwidth}
    \includegraphics[width=\textwidth]{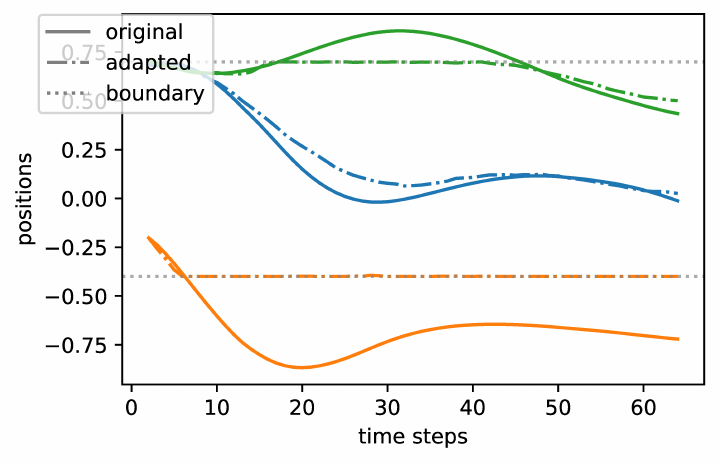}
\end{subfigure}
\vspace{-0.1cm}
\caption{Motion adaptation for position boundary. Colors indicate the axes of the Cartesian coordinate.}
\label{fig:position_bound}
\vspace{-0.2cm}
\end{figure*}

\subsection{Distances to the original trajectories}

We split the dataset into 20,000 training, 2048 validation, and 2048 test samples. A single robot was trained and then evaluated on its performance. Fig.~\ref{fig:distanceEE} and \ref{fig:distanceJoints} illustrate the generalization from the prior and the reconstruction.
Evaluation metrics include the Root Mean Square Error (RMSE) measured in degrees for joint movements, and an L2 Norm, measured in meters, for assessing the trajectories for the end-effector. Assuming endpoint distances are the same, a larger distance for trajectories from the prior distribution to the original demos is advantageous, as it demonstrates the \textit{diversity} of the learned prior. Except that, lower distance values to the original samples
indicate better performance, where the distance reflects the \textit{fidelity} of the model's output approximating the original trajectories. Rather than determining the endpoint distance, we calculate the distance at $x_{T-1}$, since the $x_T$ is given.

\subsection{Generalization on an unknown robot}

Scaling up models across various robots presents significant challenges but offers substantial benefits, as noted in \cite{padalkar2023open}. To assess our model's ability to generalize, we conducted an experiment involving different robots.
In this experiment, the training dataset comprises IIWA, UR5e, and Sawyer robots operating in Cartesian space. The validation dataset uses the Jaco robot, while the Panda robot is the test dataset. For comparison of single-robot training, the training dataset was divided into three subsets, each corresponding to a specific robot, with the same validation and test datasets used across all experiments. The results from the single-robot training are aggregated for presentation. As illustrated in Fig.~\ref{fig:MultiSingleRobot}, training across multiple robots demonstrates superior generalization to an unknown robot.

\subsection{Adaptation and generalization}

\begin{table}[!htbp]
\centering
\caption{Velocity and acceleration limits.}
\begin{tabular}{lcc}
\toprule
 methods & samples & max values  \\
\midrule
velocity bound & prior  & 0.050  \\
& posterior  & 0.050 \\
& boundary  & 0.050 \\
& original data & 0.094 \\
\midrule
acceleration bound & prior & 0.020  \\
& posterior &  0.020 \\
& boundary & 0.020 \\
& original data & 0.031 \\
\bottomrule
\label{tab:v_a_limit}
\end{tabular}
\end{table}

\begin{figure*}[!htbp]
\centering
\begin{subfigure}[b]{0.32\textwidth}
    \includegraphics[width=\textwidth]{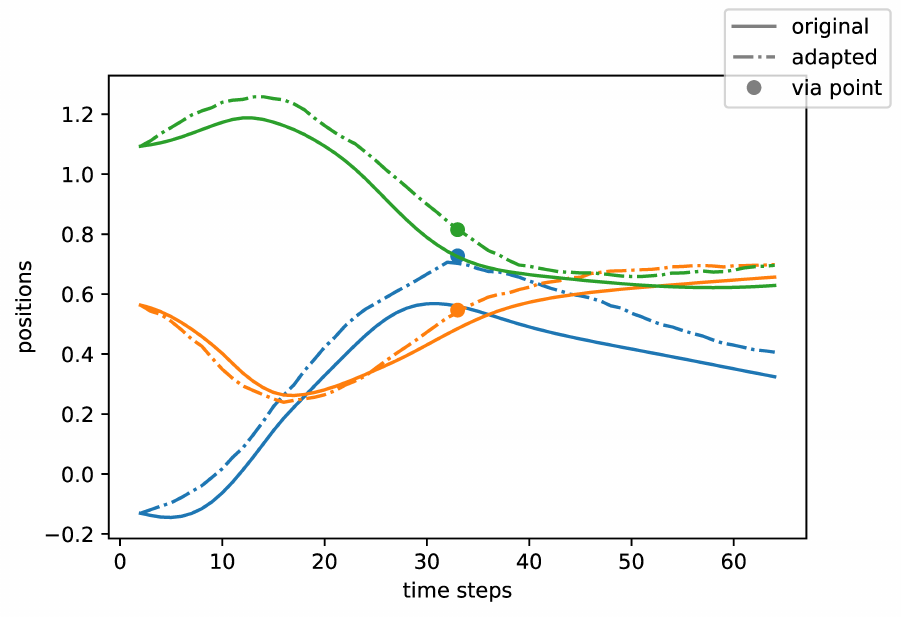}
\end{subfigure}
\hfill
\begin{subfigure}[b]{0.32\textwidth}
    \includegraphics[width=\textwidth]{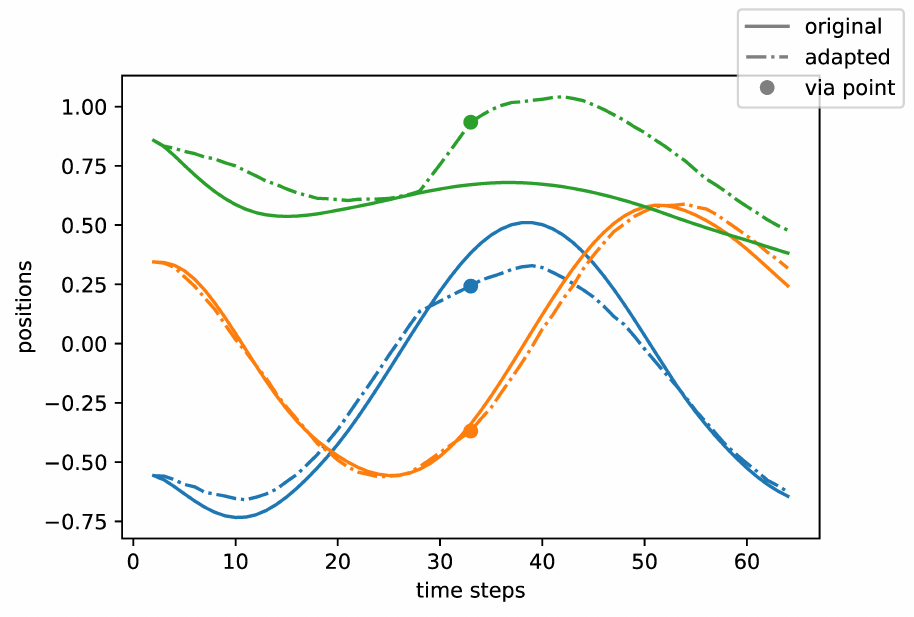}
\end{subfigure}
\hfill
\begin{subfigure}[b]{0.32\textwidth}
    \includegraphics[width=\textwidth]{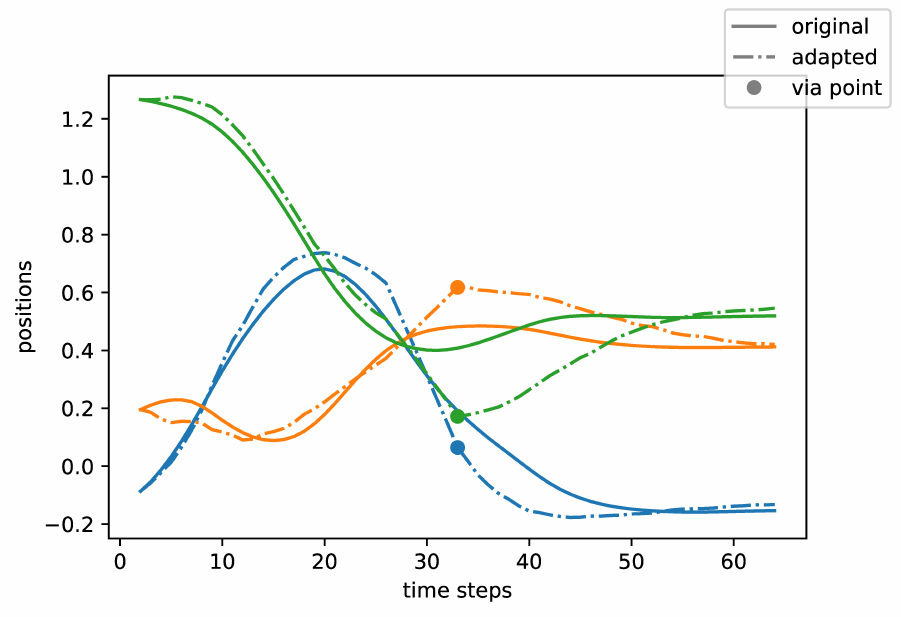}
\end{subfigure}
\\
\begin{subfigure}[b]{0.3\textwidth}
    \includegraphics[width=\textwidth]{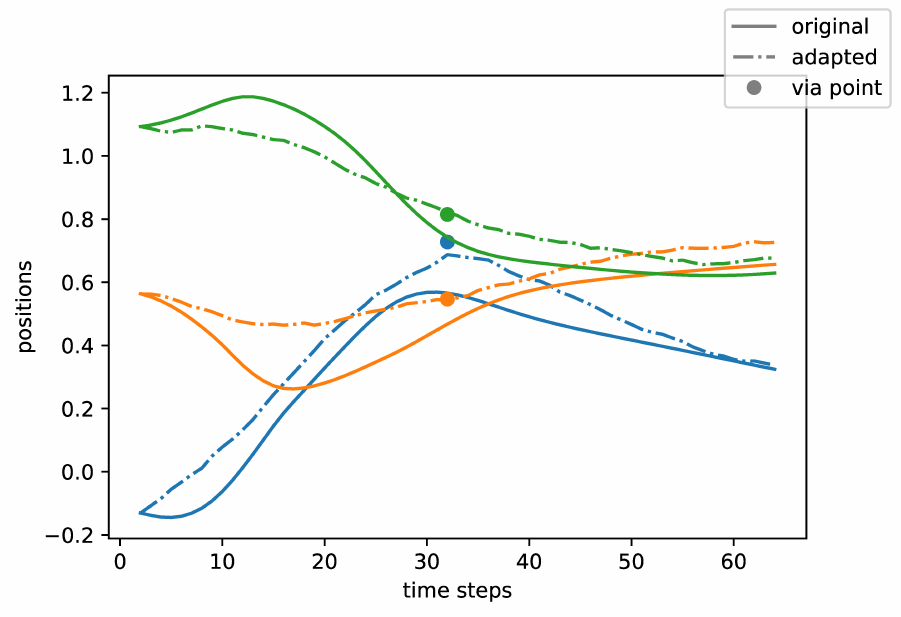}
\end{subfigure}
\hfill
\begin{subfigure}[b]{0.3\textwidth}
    \includegraphics[width=\textwidth]{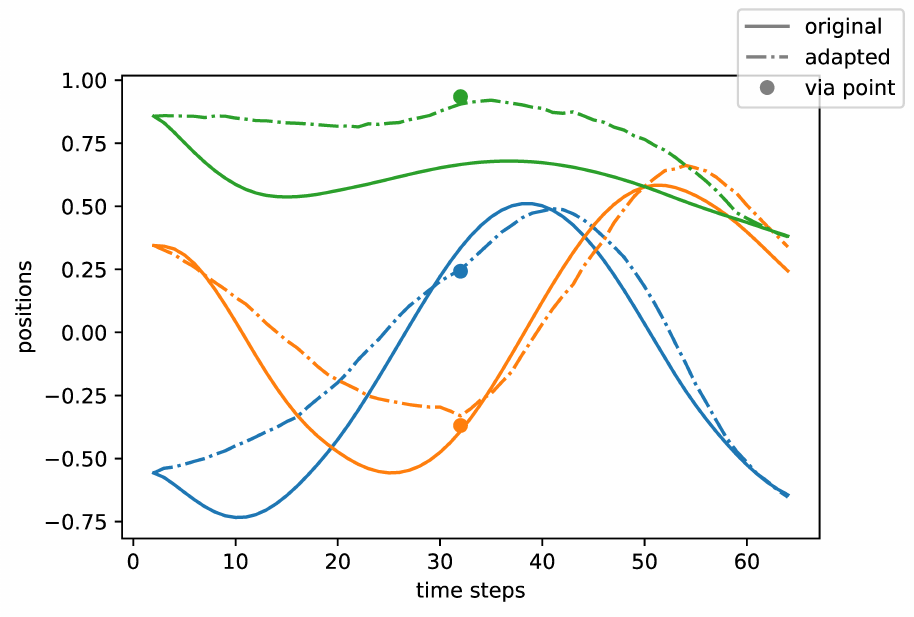}
\end{subfigure}
\hfill
\begin{subfigure}[b]{0.3\textwidth}
    \includegraphics[width=\textwidth]{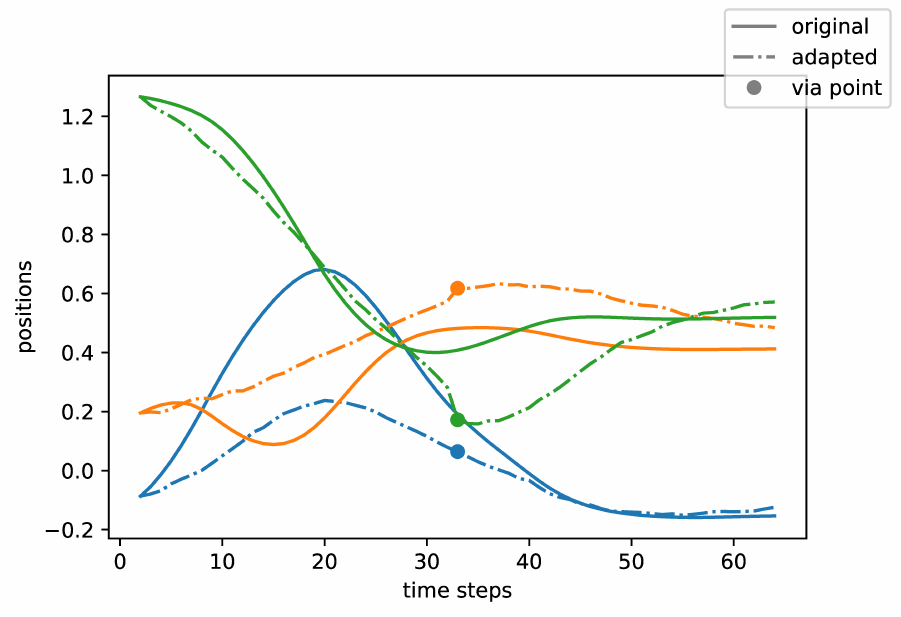}
\end{subfigure}
\vspace{-0.1cm}
\caption{Brownian bridge for via point with (upper) and without (lower) CBS. Colors indicate the axes of the Cartesian coordinate.}
\label{fig:exp_viapoint}
\vspace{-0.2cm}
\end{figure*}

\begin{figure*}[!htbp]
\centering
\begin{subfigure}[b]{0.3\textwidth}
    \includegraphics[width=\textwidth]{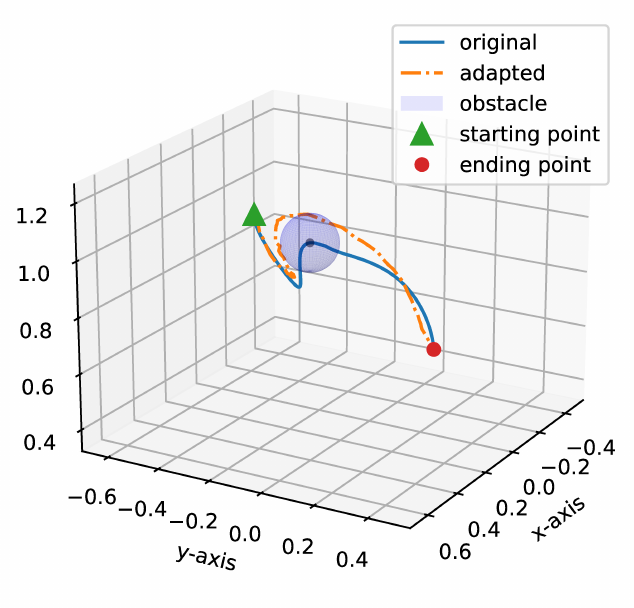}
\end{subfigure}
\hfill
\begin{subfigure}[b]{0.3\textwidth}
    \includegraphics[width=\textwidth]{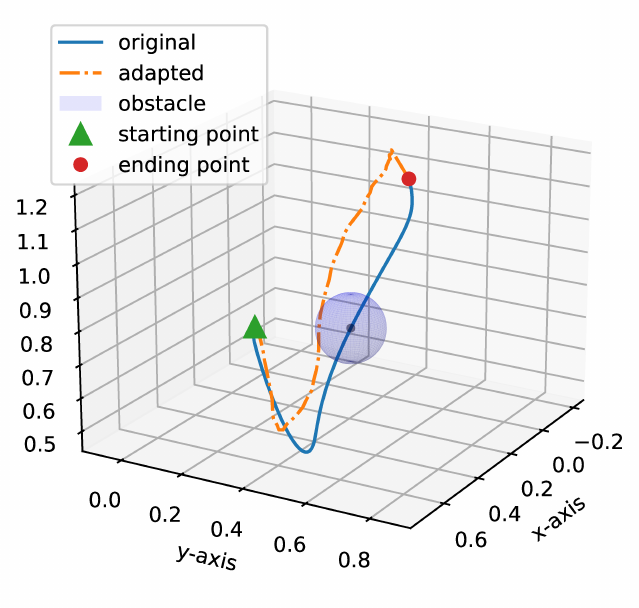}
\end{subfigure}
\hfill
\begin{subfigure}[b]{0.3\textwidth}
    \includegraphics[width=\textwidth]{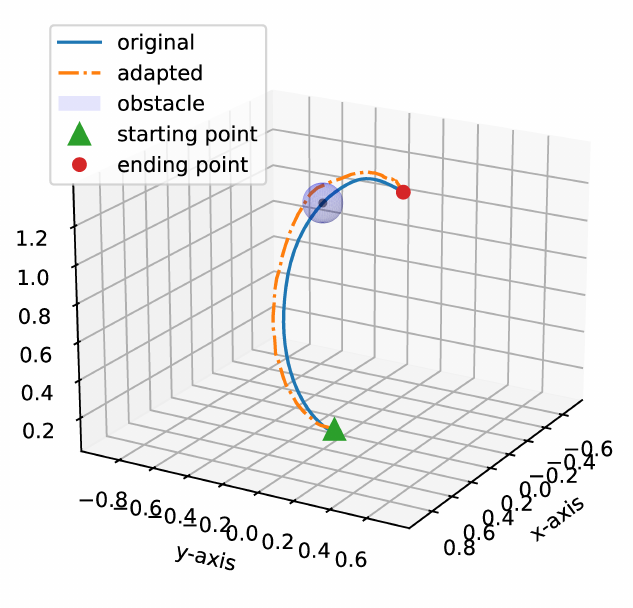}
\end{subfigure}
\vspace{-0.4cm}
\\
\begin{subfigure}[b]{0.3\textwidth}
    \includegraphics[width=\textwidth]{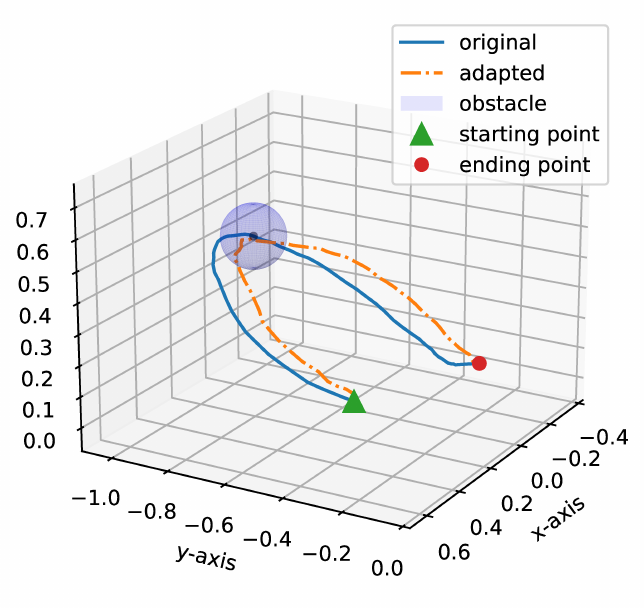}
\end{subfigure}
\hfill
\begin{subfigure}[b]{0.3\textwidth}
    \includegraphics[width=\textwidth]{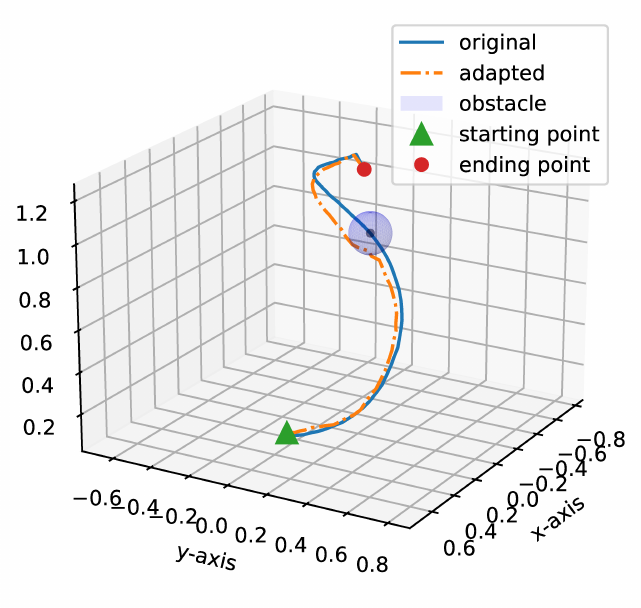}
\end{subfigure}
\hfill
\begin{subfigure}[b]{0.3\textwidth}
    \includegraphics[width=\textwidth]{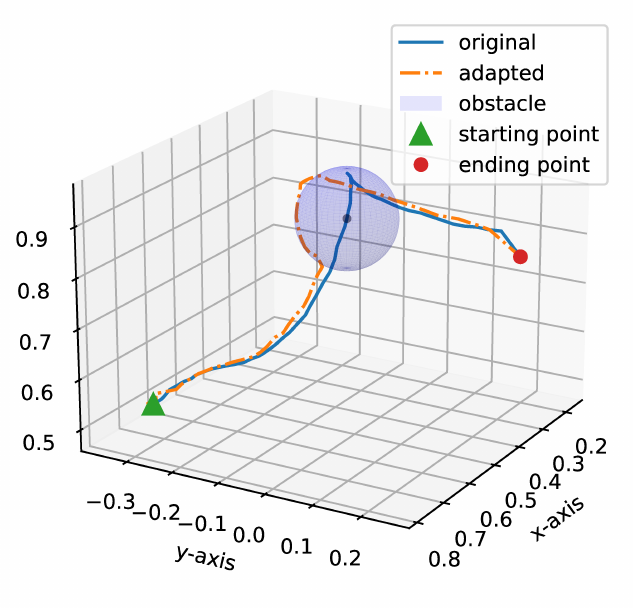}
\end{subfigure}
\vspace{-0.1cm}
\caption{Adaption for obstacle avoidance. (Upper) Obstacle avoidance with the samples from the posterior. (Lower) Obstacle avoidance with samples from the prior, where the original refers to the samples from the prior without adaptation of the decoder. All of them successfully avoid the obstacles, although the visualization is not perfect due to the viewpoint.}
\label{fig:avoidance}
\vspace{-0.2cm}
\end{figure*}

The experiments are trained on a Panda robot. Our adaption and generalization are suitable for both samples from priors and posterior. 
To enhance visualization and compare with demonstrations, we present all results based on samples from posteriors, except for those in Sections~\ref{sec:exp_obstacle} and \ref{sec:exp_bound} which also use samples from priors.

\subsubsection{Position bound}

As depicted in Figure~\ref{fig:position_bound}, the generated trajectories are supposed to follow the original demonstrations, with adaptations made to ensure compatibility within the cuboid boundaries.

\subsubsection{Velocity and acceleration bound}
\label{sec:exp_bound}

We have established boundaries for both velocity and acceleration, as detailed in Table \ref{tab:v_a_limit}. Although the table presents only a single boundary for each type of boundary, these boundaries can be set to arbitrary values.

\subsubsection{Via points}

We adapted the movements for via point (see Fig.~\ref{fig:exp_viapoint}) using the Brownian bridge.
We set the hyperparameter $\sigma_\mathrm{vp}$ in a similar range of the output  $\sigma$ from the autoregressive decoder.
The experiments show the comparisons between via points without and with CBS, where the beam search size is set to five. The findings indicate that when CBS is used, the trajectories appear more natural and closer to the original demos. On the contrary, as CBS is a historical search method, a Brownian bridge without CBS enables online trajectory generation.

\subsubsection{Obstacle avoidance}
\label{sec:exp_obstacle}

In the experiment, we used a beam search with a size of 25. The temperature of the output, which is the scale of the output STD, was set to two. Fig.~\ref{fig:avoidance} demonstrates how the adapted trajectories effectively navigate to avoid spherical obstacles for samples drawn from both the prior and posterior distributions.

\section{Conclusion}

In this paper we introduce a novel Transformer-based auto-encoder framework for robot motion generation and adaptation, leveraging Learning from Demonstration (LfD). 
The latent space of the autoencoder can be viewed a motion representation similarly as in movement primitives.
The conditioning of the trajectory generation on goal an motion representation is implemented through flexible cross-attention mechanisms. 
By using an autoregressive decoder and online (approximate) inference methods our approach generates trajectories that adapt to various task space constrains such as obstacle avoidance, via-points, and various position, velocity, and acceleration bounds.
The experiments validated the framework's robustness and adaptability across different robots and various trajectory constraints.

\newpage
\clearpage

\bibliography{mybib}
\bibliographystyle{IEEEtran}

\appendix

\subsection{Training}
\label{sec:app}

Training VAEs and CVAEs can be challenging due to local minima, posterior collapse or over-regularization \cite{sonderby2016,bowman2015generating,kingma2016improved}. Sequence-to-sequence models with strong decoders are often prone to posterior collapse. The authors in \cite{roberts2018hierarchical} reduced the capability of the decoder using the hierarchical decoder. Alternatively, reformulating the objective or using specialized training strategies can also alleviate some optimization problems.
The authors in 
\cite{rezende2018taming} reformulate training as a constrained optimization problem where the reconstruction term $ -\mathrm{E}_{\hat{p}(x_{1:T})}
    \mathrm{E}_{q_{\phi}(z;x_{1:T})}[\log p_{\theta}(x_{2:T-1} \vert x_{1,T}, z)]$ term is constrained to achieve specific reconstruction accuracy.
The authors in \cite{klushyn2019learning} train a hierarchical VAE model using a similar reformulation and use a scheduled pre-training of the reconstruction term.  
To attain a required reconstruction loss (powerful decoder) and to also avoid posterior collapse we formulate the optimization problem as  
\begin{align*}
    \min_{\theta, \phi} & \: 1
    \\
    \mathrm{s.t.} & \:
    \mathrm{E}_{\hat{p}(x_{1:T})}
    \mathrm{E}_{q_{\phi}(z;x_{1:T})}[-\log p_{\theta}(x_{2:T-1} \vert x_{1,T}, z)] \leq \xi_{\mathrm{rec}}
    \\
    & \:
    \mathrm{E}_{\hat{p}(x_{1:T})}
    \mathrm{KL}[q_{\phi}(z;x_{1:T}) \,\|\,
    p_{\theta}(z;x_{1,T})
    ] \leq\xi_{\mathrm{kl}}.
\end{align*}
The resulting Lagrangian will have two multipliers
$\lambda_{\mathrm{rec}}$ and $\lambda_{\mathrm{kl}}$ that adaptively re-weight the reconstruction and the KL (regularization) terms. We use the exponential method of multipliers \cite{bertsekas}, that is, we apply the update
$\lambda_{\mathrm{kl}}^{(k+1)} = \lambda_{\mathrm{kl}}^{(k)} \cdot \exp \{
   \eta (    - \mathrm{E}_{\hat{p}(x_{1:T})}
    \mathrm{KL}[q_{\phi}(z;x_{1:T}) \,\|\,
    p_{\theta}(z;x_{1,T})
    ] +\xi_{\mathrm{kl}})
\}$ and a corresponding one for $\lambda_{\mathrm{rec}}$.

\subsection{Architecture and computation}

The computational analyses of this research used an NVIDIA GeForce GTX 1080 Ti GPU, with PyTorch version 2.1.0 for implementation.

The model's encoder depth was set to 3, with an embedding dimension of 16, and a latent vector length of 32. The decoder was configured with 8 heads and 10 layers. For optimization, the RAdam optimizer \cite{liu2019variance}, was chosen, configured with beta values of 0.9 and 0.999, and a learning rate of 0.0005.
Batch sizes were 256 for the data in Cartesian space and 128 for the joint space. Mixup \cite{zhang2017mixup} was used as a data augmentation for the joint space data.

\end{document}